\documentclass{article}

\usepackage[final]{neurips_2024}

\usepackage[utf8]{inputenc} 
\usepackage[T1]{fontenc}    
\usepackage{hyperref}       
\usepackage{url}            
\usepackage{booktabs}       
\usepackage{amsfonts}       
\usepackage{nicefrac}       
\usepackage{microtype}      

\usepackage{amsmath}
\usepackage{amssymb}
\usepackage{mathtools}
\usepackage{algorithmic}
\usepackage{multirow}
\usepackage{amssymb}
\usepackage{graphicx}
\usepackage[ruled,vlined,linesnumbered]{algorithm2e}
\SetKwInput{Input}{Input}
\SetKwInput{Output}{Output}
\SetKw{To}{ to }
\SetKw{Print}{Print }
\SetKwFor{FOR}{for}{do}{endFor}
\usepackage{cleveref}
\usepackage{subcaption}
\usepackage{diagbox}

\usepackage{tabularx}
\usepackage[table]{xcolor}
\usepackage{wrapfig, booktabs}

\usepackage[inkscapelatex=false]{svg}

\usepackage{pifont}
\newcommand{\xmark}{\ding{55}}

\DeclareMathOperator*{\argmax}{arg\,max}
\DeclareMathOperator*{\argmin}{arg\,min}
\DeclareMathOperator{\EX}{\mathbb{E}}

\title{Color-Oriented Redundancy Reduction in Dataset Distillation}

\author{%
\textbf{Bowen Yuan} \quad Zijian Wang \quad Mahsa Baktashmotlagh \quad \textbf{Yadan Luo} \quad \textbf{Zi Huang} \\
\texttt{\{bowen.yuan, zijian.wang, m.baktashmotlagh, y.luo, helen.huang\}@uq.edu.au} \\
The University of Queensland \\
}

\begin{document}

\maketitle

\begin{abstract}


Dataset Distillation (DD) is designed to generate condensed representations of extensive image datasets, enhancing training efficiency. Despite recent advances, there remains considerable potential for improvement, particularly in addressing the notable redundancy within the color space of distilled images. In this paper, we propose AutoPalette, a framework that minimizes color redundancy at the individual image and overall dataset levels, respectively. 
At the image level, we employ a palette network, a specialized neural network, to dynamically allocate colors from a reduced color space to each pixel. The palette network identifies essential areas in synthetic images for model training and consequently assigns more unique colors to them. At the dataset level, we develop a color-guided initialization strategy to minimize redundancy among images. Representative images with the least replicated color patterns are selected based on the information gain.
A comprehensive performance study involving various datasets and evaluation scenarios is conducted, demonstrating the superior performance of our proposed color-aware DD compared to existing DD methods. The code is available at \url{https://github.com/KeViNYuAn0314/AutoPalette}.

\end{abstract}

\section{Introduction}

Large-scale training data is essential for achieving high model performance. However, the sheer volume of the data poses significant challenges, including computational inefficiency, prolonged training times, and substantial storage overhead. \textit{Data Distillation} (DD) \cite{wang2018dataset} offers a promising solution to this problem. By synthesizing a smaller dataset from the original dataset, DD allows models trained on the distilled dataset to attain comparable performance to those trained on the full dataset, thereby reducing the resources needed for training.

Existing DD primarily minimizes the difference between the network trained on the full dataset and the network trained on the synthetic dataset. Different surrogate functions have been implemented to quantify such differences, including performance matching \cite{wang2018dataset, nguyen2021kipimprovedresults}, feature distribution matching \cite{wang2022cafe, zhao2023dataset} and model gradient matching \cite{zhao2020dataset, cazenavette2022dataset}. Generally speaking, DD considers the synthetic images as parameters and directly optimizes them.
Building on this concept, parameterization-based dataset distillation (PDD) extends DD by enhancing the storage utility and reducing redundancy in the image space. Parameterization-based DD methods represent the synthetic dataset in a lower-dimensional space and then, reconstruct synthetic images for model training. Current parameterization-based DD includes: learning in a spatially down-sampled space \cite{kim2022dataset}, factorizing distilled images ~\cite{deng2022remember, liu2022dataset}, optimizing latent embeddings and generators \cite{zhao2022synthesizing, cazenavette2023glad}, and selecting informative frequency bands \cite{shin2024frequency}.

While the existing PDD methods have shown promising results, most of the methods overlook the redundancy in the color space, thereby falling short of achieving optimal parameterization performance. We argue that reducing the number of unique colors within one image can have \textit{minimal} impact on the low-level discriminative features (\textit{e.g.}, shapes, edges) required for model training. Moreover, images within the same class typically share similar color distributions; therefore, dedicating storage to \textit{unique} class patterns rather than storing the replicated color information would be more cost-effective.

To address the limitations of existing PDD approaches, we propose a color-oriented redundancy reduction framework, namely AutoPalette. Specifically, AutoPalette contains an efficient plug-and-play palette network to tackle the issue of color space redundancy within one image. This palette network transforms 8-bit color images into representations with fewer colors (\textit{e.g.}, 4-bit) by aggregating pixel-level color information from input images. To enhance the color utility, we design two additional losses on top of the dataset distillation loss: the maximum color loss and the palette balance loss. The maximum color loss ensures that each color in the reduced color space is allocated to at least one pixel, while the palette balance loss balances the number of pixels allocated to each color. The palette network synthesizes image datasets with a reduced color space while preserving essential features of the original images. Furthermore, we equip AutoPalette with a color-guided initialization module to suppress the redundancy in-between synthetic images. The module selects the samples with low replication after color condensation as the synthetic set initialization, whereas information gain is adopted to quantify replication.

\noindent\textbf{Contributions.} We propose AutoPalette, a color-oriented redundancy reduction framework for data distillation tasks, enhancing storage efficiency by reducing the number of colors in the images while preserving essential features; We seamlessly equip the distillation framework with a guided initialization strategy that selects images with diverse structures in the reduced color space for initialization; Extensive experimental results on three benchmark datasets show that the model trained on the 4-bit images synthesized by our framework achieve competitive results compared to the models trained on 8-bit images synthesized by other DD methods. With the same storage budget, our method outperforms others by $1.7\%$, $4.2\%$ on CIFAR10 and CIFAR100.

\begin{figure}[!t]
\centering
    \includegraphics[width=0.9\linewidth]{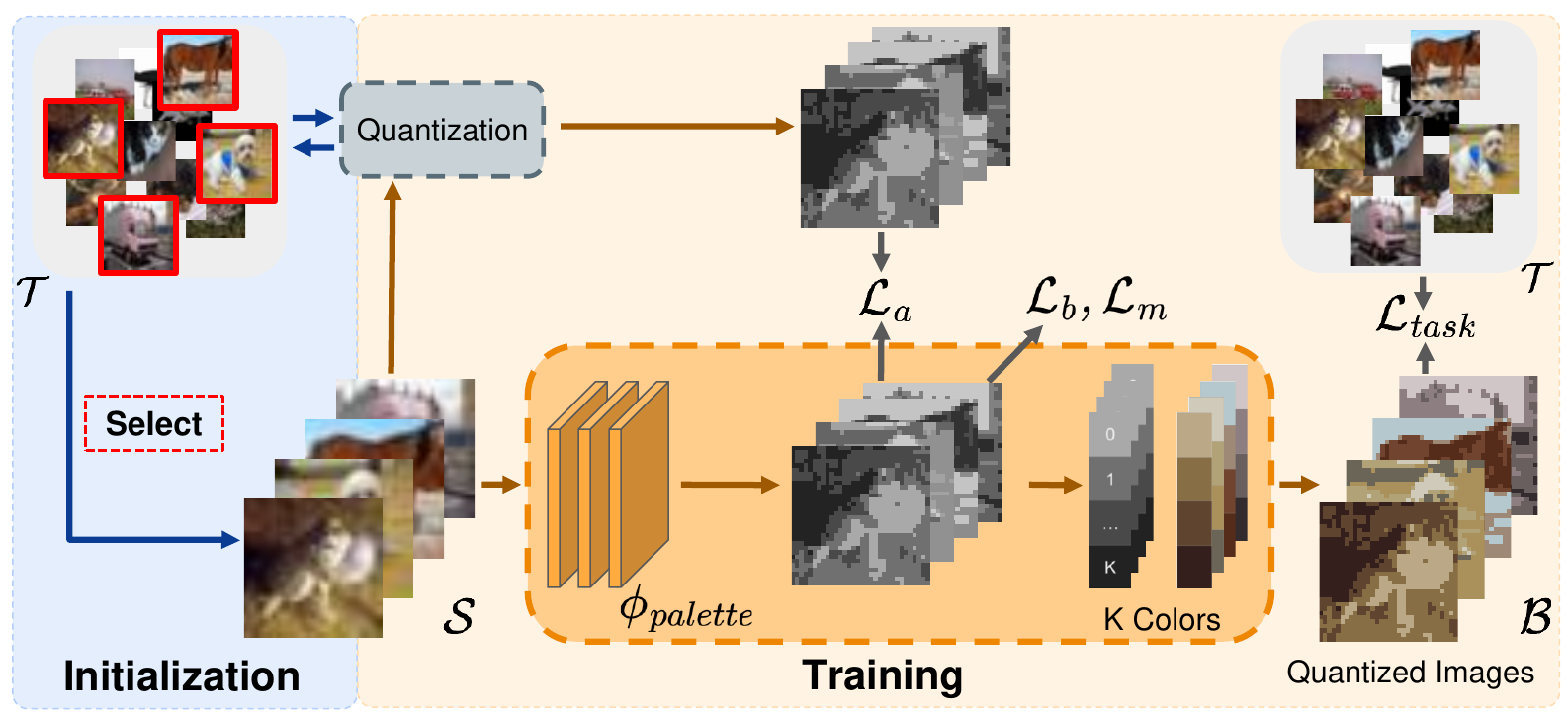}
    \caption{The overview of the proposed AutoPalette framework. Initialization: We compare the information gain of quantized images to select the images used in the initialization stage. Training: We forward the synthetic data to the palette network to obtain the color-reduced images. The objective functions of palette network include $\mathcal{L}_a$, $\mathcal{L}_b$, $\mathcal{L}_m$ and $\mathcal{L}_{task}$. The synthetic dataset is updated by solely optimizes $\mathcal{L}_{task}$ .}
\label{fig:flowchart}
\vspace{-1em}
\end{figure}

\section{Related Work}
\subsection{Dataset Distillation} 
Dataset distillation aims to synthesize a small but informative dataset, enabling models trained on these synthetic data to achieve comparable performance to those trained on the complete dataset. Wang \textit{el al.} \cite{wang2018dataset} firstly proposed a meta-model learning approach to optimize a synthetic dataset matching the model performance of large scale dataset. Early works adopt performance matching frameworks \cite{nguyen2020dataset, nguyen2021dataset, zhou2022dataset, vicol2022implicit}, and optimize synthetic data using model performance rolling over the training process on the original dataset. Distribution matching methods \cite{wang2022cafe, zhao2023dataset, zhao2023improved, sajedi2023datadam} address high complexity issues in bi-level optimization by matching one step feature distributions between synthetic data and original real data. Gradient matching \cite{zhao2020dataset, zhao2021dataset, liu2023dream} and trajectory matching \cite{cazenavette2022dataset, cui2023scaling, guo2023towards, du2023minimizing} approaches aim to match model parameters' gradients for single or multiple training steps, leading networks trained on synthetic data and original data to follow similar gradient descent trajectories. 

\subsection{Parameterization-based Dataset Distillation} 
Apart from finding matching objectives between the synthetic dataset and the original full dataset, another aspect of data distillation involves appropriately parameterizing synthetic data in different yet more efficient representatives in memory space. Without storing synthetic data as individual spatial representations, parameterization comprehends mutual characteristics between data instances and regenerates more data instances of the original input representations. IDC \cite{kim2022dataset} stores images in a low-resolution manner to conserve storage resources, and upsamples to the original scale for usage. Factorization methods conjecture inter-class data share mutual and independent information and generate synthetic data based on combinations of bases. Bases can be either spatial representations \cite{deng2022remember}, frequency domains \cite{wei2024sparse}, or embeddings decoded by networks \cite{lee2022dataset, liu2022dataset, wei2024sparse, zhao2022synthesizing, cazenavette2023generalizing, wang2023dim}.

\subsection{Color Quantization}
Color quantization \cite{orchard1991color, deng1999peer, achanta2012slic, wu1992color} intends to aggregate similar colors and transform them using one representative color. Accordingly, images with a reduced color palette require less storage as the pixel values can be encoded in fewer bits. To uphold optimal image authenticity, traditional color quantization methods such as Median Cut \cite{heckbert1982color}, dithering \cite{dithering}, and OCTree \cite{gervautz1988simple} typically employ color quantization as a color clustering problem. They commonly devise strategies to identify similar or neighboring colors for quantization purposes. On the other hand, parameter-based methods \cite{van2016pixel, mentzer2019practical, hou2020learning, hou2022learningstructureimagecolors} not only rely on predefined heuristics but also leverage neural networks to learn patterns and relationships to compress images to lower bits.


\begin{figure*}[t!]
    \centering
        \centering
        \begin{subfigure}[t]{0.265\textwidth}
            \centering
            \includegraphics[width=\textwidth]{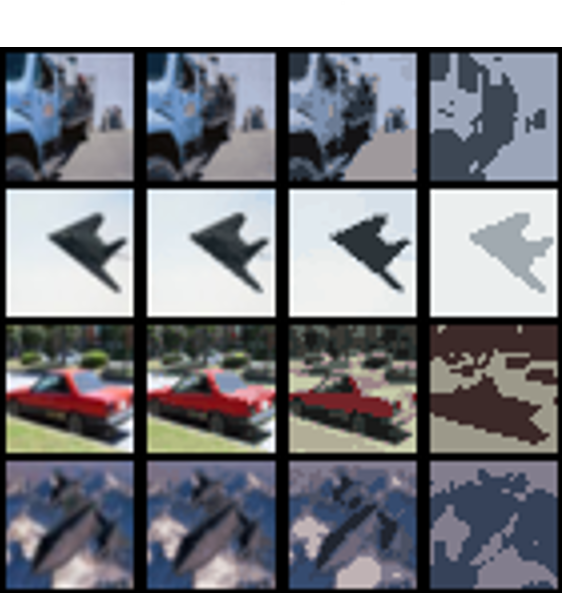}
            \caption{}
        \end{subfigure}
        \hfill
        \begin{subfigure}[t]{0.33\textwidth}
            \centering
            \includegraphics[width=\textwidth]{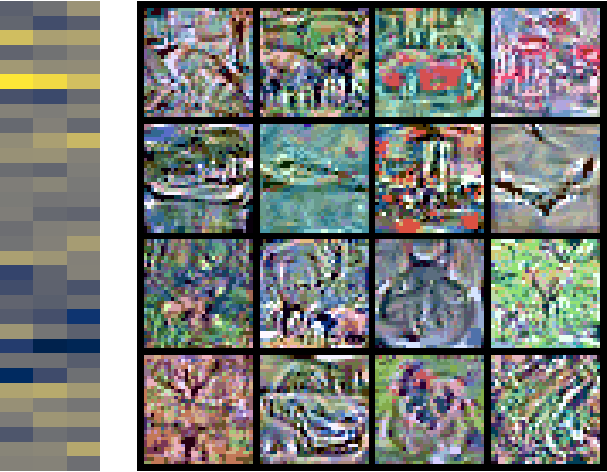}
            \caption{}
        \end{subfigure}
        \hfill
        \begin{subfigure}[t]{0.33\textwidth}
            \centering
            \includegraphics[width=\textwidth]{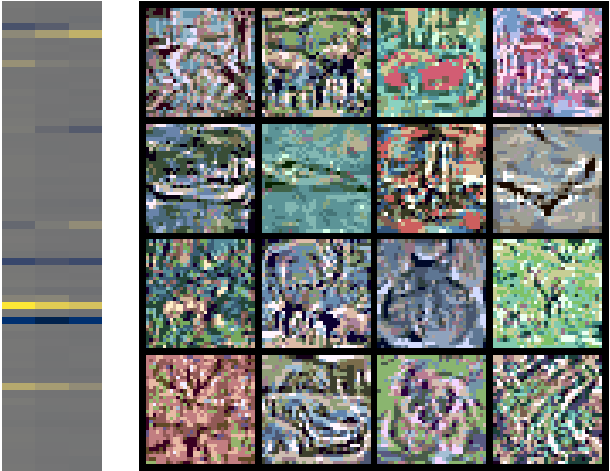}
            \caption{}
        \end{subfigure}
        \hfill
        \caption{The visualization of (a) images under 8, 6, 3, 1-bit color depths (b-c) color condensed synthetic images and their color palette. (b) our full model (c) our full model without palette loss. The larger difference among rows of a color palette indicates better color utilization.}\label{fig:even_bucket}
    %
\vspace{-1.em}
\end{figure*}

\section{Methodology}

\subsection{Notations and Preliminary}
Dataset distillation aims to learn a small but representative synthetic dataset $\mathcal{S} = {\{({\Tilde{x}}^{i}, {\Tilde{y}}^{i})\}}^{|\mathcal{S}|}_{i=0}$ from a given large scale dataset $\mathcal{T} = {\{({x}^{i}, {y}^{i})\}}^{|\mathcal{T}|}_{i=0}$. Here, $|\mathcal{S}|$ and $|\mathcal{T}|$ denote the number of samples in the synthetic dataset and original large dataset, where $|\mathcal{S}|\ll|\mathcal{T}|$. By training on the synthetic dataset $\mathcal{S}$, a model $\phi(\cdot;\theta)$ is aimed to achieve performance comparable to that of a model trained on the original dataset $\mathcal{T}$. The objective of dataset distillation can be formulated as a bi-level optimization problem:
\begin{equation}
\label{bilevel eq}
    \underset{\mathcal{S}}{\min}\EX_{\theta}[\mathcal{L}(\mathcal{T}, \theta_{\mathcal{S}})] \text{, where }  \theta_{\mathcal{S}} = \underset{\theta}{\argmin}\mathcal{L}(\mathcal{S}, \theta),
\end{equation}

where $\mathcal{L}(\cdot, \cdot)$ and $\theta$ represent loss function and parameters of the networks, respectively. The inner-loop optimizes the network on the synthetic dataset and the outer loop evaluates the trained network on the real dataset.

The bi-level meta learning in Eq. \ref{bilevel eq} requires inner-loop training during every training steps, and thus suffers from inevitable computational cost. Therefore, some of the existing methods \cite{zhao2020dataset, cazenavette2022dataset, zhao2023dataset} try to avoid unrolled back-propagation in the inner loop using various surrogate objectives, thereby the optimal synthetic dataset $\mathcal{S}^*$ can be obtained by optimizing:
\begin{equation}
    \mathcal{S}^{*} = \underset{\mathcal{S}}{\argmin}\EX_{\theta}[\mathcal{L}(\phi(\mathcal{T}; \theta), \phi(\mathcal{S}; \theta))].
\end{equation}



In parameterization methods for dataset distillation, the synthetic dataset $\mathcal{S}$ is stored in more efficient representations consisting of bases $\mathcal{B}\in \mathbb{R}^{N \times C \times H \times W}$ and a set of transformation functions $F: \mathcal{B} \rightarrow \mathcal{S}$, which generate the synthetic dataset. Here, $N$ denotes the number of bases, $C$ represents the channel, and $H$ and $W$ are the height and width of the bases, respectively.  In this paper, we propose a color transformation that reduces the number of unique colors in the image bases while preserving the essential details after color reduction for model training.

\subsection{Overview}
In this paper, we explore a new dimension of parameterization-based dataset distillation, concentrating on optimizing storage efficiency by minimizing color space redundancy. We argue that complex color representation within storage-sensitive distilled images is not crucial for training networks. Instead, the limited storage budget for distilled images should be allocated to novel samples that exhibit diverse object structures. The overall framework of AutoPalette is illustrated in Figure \ref{fig:flowchart}.

The proposed AutoPalette framework for dataset distillation consists of two components, including a palette network and a color-guided initialization strategy. The palette network is designed to enhance the color utility of the synthetic images in reduced color space by generating pixel-level color mappings. Accordingly, the original images $\Tilde{x} \in \mathbb{R}^{C \times H \times W} \in \mathcal{S}$ are transformed into color-condensed images $\mathbf{b} \in \mathbb{R}^{C \times H \times W} \in \mathcal{B}$ with a reduced color spectrum, where we denote the process as $\phi_{palette}: \mathcal{S} \rightarrow \mathcal{B}$. RGB image dataset generally contains 256 distinct colors for each channel, and the palette network aims to reduce the number of colors by generating a color palette containing only $K$ colors. As such, the synthetic dataset can be stored in a low-bit format, rather than the conventional 8-bit format. To better leverage the diverse color feature information within the original dataset, we equip the proposed framework with a novel initialization strategy, which employs the generalized graph cut function to select representative images for initialization. Our method dynamically evaluates the impact of real images based on their color structures and initializes the synthetic dataset with the images that yield the highest information gains of graph cut functions. The subsequent sections will illustrate each module in detail.

\subsection{Color Reduction via Palette Network.}
\label{color Palette method}
The core of our proposed parameterization method for dataset distillation is to condense the number of unique colors in an image so that the image can be stored with a more efficient manner. One of the key challenges of reducing the unique color number in dataset distillation lies in local discriminative feature preservation. When an image is represented with a smaller range of color, it is inevitable for some pixels to be merged to their neighbour color blocks. In this case, some of the local discriminative features (\textit{e.g.}, edge, shape, \textit{etc.}) can be erased or distorted in the color-reduced images, hindering the network trained on them to achieve the optimal performance. 

To alleviate this issue, we design a simple yet effective network, namely palette network, which learns the pixel-level color allocation in the pruned color space with the discriminative feature maximally preserved. Particularly, the color palette network predicts the probability map $\mathbf{m} \in \mathbb{R}^{C \times H \times W \times K}$, which indicates the probability of a pixel being allocated to a color of the $K$-dimensional reduced color space by forwarding an image $\Tilde{x}$ to the palette network:
\begin{equation}
    \mathbf{m} = \phi_{color}(\Tilde{x}; \theta_{c}),
\end{equation}
where $\theta_c$ is the parameters of the palette network.

Given a base image $\mathbf{b}$ and its corresponding probability map $\mathbf{m}$, we formulate the palette $\Tilde{\mathbf{m}} \in \mathbb{R}^{C \times K}$ as the average pixel values of all pixels assigned to the same color buckets index:
\begin{equation}
\label{eq2}
    \Tilde{\mathbf{m}}_{c, k} = \frac{\sum_{c, i, j} \Tilde{x}_{c, i, j} \cdot \delta^{\mathbf{m}}_{c,i,j}(k)}{\sum_{c, i, j} \delta^{\mathbf{m}}_{c,i,j}(k)},
\end{equation}
where $c$ and $k$ denote the $c$-th channel for the $k$-th quantized color, i and j denote the vertical and horizontal pixel position in an image, and $\delta^{\mathbf{m}}$ is the Kronecker delta function, which equals 1 if $\argmax\mathbf{m}_{c, i,j}$ equals $k$. 




Once the color palette $\Tilde{\mathbf{m}}$  and a probability map $\mathbf{m}$ are generated for an image, its color condensed image $\mathbf{b}$, with the number of unique colors per channel reduced to $K$, can be generated by an index searching process:
\begin{equation}
\label{eq3}
    \mathbf{b}_{c,i,j} = \Tilde{\mathbf{m}}[c, \mathbf{h}], \text{ where } \mathbf{h} = \argmax_k\mathbf{m}_{c,i,j}.
\end{equation}






To learn the palette network, a straightforward solution is to optimize the distillation task loss during the training stage. However, we can observe from Fig. \ref{fig:even_bucket} that solely relying on the task loss for training the palette network leads to most of the palette being inactivated. The network tends to assign pixels to a limited number of color buckets, which strongly limits the capacity and expressiveness of the distilled images. Therefore, two additional losses, namely \textbf{maximum color loss} and \textbf{palette balance loss}, are incorporated to enhance the utility of color buckets in synthetic images. In particular, maximum color loss, $\mathcal{L}_{m}$ encourages the palette network to generate color allocation such that each color bucket is at least filled with one pixel within color palette. By aggregating the maximum confidences from probability index maps across the spatial dimensions, we define the maximum color loss as:
\begin{equation}
    \mathcal{L}_{m} = - \frac{1}{CK} \sum_{c=1}^{C} \sum_{k=1}^{K} \max_{(h, w)}(\mathbf{m}_{c, h, w, k}),
\end{equation}

While the maximum color loss ensures the activation of each color bucket in the palette, the distribution of pixel numbers in color buckets can still be extremely imbalanced. Therefore, our framework leverages a palette balance loss $\mathcal{L}_{b}$, which encourages a more balanced usage of the buckets within the color palette by promoting color-wise entropy. We formulate the palette balance loss as the  entropy of $\mathbf{m}$ over the spatial dimensions:
\begin{equation}
    \mathcal{L}_{b} = \frac{1}{CK} \sum_{c=1}^{C}  \sum_{k=1}^{K} P(\frac{1}{HW}\sum_{i=1}^{H} \sum_{j=1}^{W} \mathbf{m}_{c,i,j,k}) \log P(\frac{1}{HW}\sum_{i=1}^{H} \sum_{j=1}^{W} \mathbf{m}_{c,i,j,k}),
\end{equation}
where $P(\cdot)$ represents the softmax function of $\mathbf{m}$ over the spatial dimensions.

By integrating the palette network with the complementary losses, we obtain the color condensed images while preserving the informative features.

\subsection{Color Guided Initialization Module}
The empirical results of previous studies have shown a strong correlation \cite{cazenavette2022dataset, zhao2023dataset} between the original images selected during initialization and the resulting distilled images in terms of visual appearance. In light of this finding, we propose an initialization method aimed at solving the redundancy problem in color-condensed synthetic images. However, since we do not have access to the optimized palette network, it is prohibitive to directly measure the information overlap within a class after color condensation. To mitigate this issue, we propose to leverage the traditional color quantization approach \cite{heckbert1982color} to approximate the output of the palette network. Here, we denote the quantized full dataset as $\mathcal{T}^Q$. Our proposed initialization strategy leverages conditional gain within submodular information theoretics to identify the most diverse images of each class after color condensation. Specifically, the conditional gain $G(\mathcal{A}|\mathcal{C})$ implies the gain of information by adding set $\mathcal{C}$ to set $\mathcal{A}$, where $\mathcal{A}, \mathcal{C} \subset \mathcal{T}^Q$ and $\mathcal{A} \cap \mathcal{C} = \emptyset$. Formally, we have:
\begin{equation}
\label{eq: conditional gain}
    G(\mathcal{A} | \mathcal{C}) = G(\mathcal{T}^Q) - G(\mathcal{C}),
\end{equation}
where $G(\cdot)$ denotes a submodular function. Submodular information functions \cite{iyer2021submodular} describe a set of combinatorial functions that satisfy the Shannon inequality \cite{shannon2001mathematical, mcgill1954multivariate} and can effectively model the diversity of a subset. In our implementation, we adopt a monotone submodular function, namely generalized graph cut~\cite{boykov2006graph}, which maximizes the similarities between samples in $\mathcal{A}$ and $\mathcal{C}$ and minimizes and dissimilarities among the samples in $\mathcal{A}$. The generalized graph cut function $G^*(\cdot)$ is defined as follows:
\begin{equation}
\label{eq: graph cut}
    G^*(\mathcal{A}|\mathcal{C}) = \sum_{i \in \mathcal{C}}\sum_{j \in \mathcal{A}}Sim(i, j) - \sum_{j_{1},j_{2} \in \mathcal{A}}Sim(j_{1}, j_{2}),
\end{equation}
where 
$i$ and $j$ denote data samples from the sets $\mathcal{C}$ and $\mathcal{A}$, respectively. $Sim(\cdot,\cdot)$ is a similarity function between two samples. Instead of directly measuring the feature level similarity, we propose to measure the similarity between the last layer gradients $\nabla\theta$ as follows: 
\begin{equation}
\label{eq:cosine kernel}
    Sim(i, j) = cos(\nabla_{\theta}\mathcal{L}_{CE}(Q(\Tilde{x}^{i}), \theta), \nabla_{\theta}\mathcal{L}_{CE}(Q(\Tilde{x}^{j}), \theta)),
\end{equation}
    where $cos(\cdot, \cdot)$ is the cosine similarity function, $Q$ denotes the Median Cut quantization method, $\Tilde{x}^{i}$ and $\Tilde{x}^{j}$ are the ith and jth samples of set $\mathcal{T}$, and $\nabla_{\theta}$ is the gradient of cross-entropy loss between the prediction and the ground truth label on the last layer of the network. By substituting Eq.\eqref{eq: graph cut} into Eq. \eqref{eq: conditional gain}, we select a representative sample for inclusion in $\mathcal{A}$ by:
\begin{equation}
\label{eq: selection criteria}
   \argmax_{c} G^*(\mathcal{A}) - 2\sum_{i \in \mathcal{A}}\sum_{c \in \mathcal{C}}Sim(i, c).
\end{equation}  
The proof for the graph cut conditional gain is provided in Appendix \ref{proof:graph cut}. From Eq. \eqref{eq: selection criteria}, we can see that the data sample obtaining the highest conditional gain may be selected. Intuitively, we select the most representative sample from the unselected set $\mathcal{C}$, whilst ensuring it is dissimilar to the already selected samples in $\mathcal{A}$. The entire color diversity selection process is provided in Algorithm \ref{algorihtm: submodular}.

By far, our initialization method can select diverse and representative samples of each class in the approximated quantization set. To minimize the difference between the approximation set and the output of the palette network, we put forward a regularization term $\mathcal{L}_{a}$. The regularization term not only constrains the color allocation shifting of palette network, but also enhances allocation consistency, so similar colors are grouped together with higher fidelity. The regularization term $\mathcal{L}_{a}$ is defined as: 

\begin{equation}
    \mathcal{L}_{a} = \frac{1}{N}{\| {\mathbf{h}} \odot {\mathbf{h}}^{\intercal} - \mathbf{h}' \odot {\mathbf{h}'}^{\intercal} \|}_{2}^{2},
\end{equation}
where 
$\mathbf{h}'$ denotes the the $\argmax$ of the color mapping indices by Median Cut over the color space, and $\odot$ denotes the element-wise multiplication resulting in a self correlation matrix for palette bucket allocations of the palette network and Median Cut. When creating index mappings for pixels, different methods might cluster the same pixels into the same group, but the indices may not match. By utilizing $\mathcal{L}_{a}$, we emphasize clustering resemblance, disregarding the order of clustering indices.

\subsection{Overall Dataset Distillation Objective}
Our framework aims to create a color-condensed synthetic version of the original dataset while maximally preserving task-related information. In line with other parameterization-based methods, we incorporate the dataset distillation loss $\mathcal{L}_{task}$ from existing works into our framework.

As such, to update the palette network, we have the overall loss function defined as:
\begin{equation}
    \argmin_{\theta_c}\mathcal{L}_{palette} = \mathcal{L}_{task} + \alpha \mathcal{L}_{m} + \beta \mathcal{L}_{b} + \gamma \mathcal{L}_{a},
\end{equation}
where $\alpha, \beta, \gamma$ are the coefficients as the weights of loss components. The synthetic set $\mathcal{S}$ is optimized as follows:

\begin{equation}
    \argmin_{\mathcal{S}}\mathcal{L}_{task} = \mathcal{L}(\phi(\mathcal{T}; \theta), \phi(\mathcal{B}; \theta)), \text{where } \mathcal{B} = \phi_{palette}(\mathcal{S}; \theta_c),
\end{equation}
where $\phi_{palette}(\cdot;\theta_c)$ denotes the color quantization process using the palette network.

\subsection{Storage Analysis}
In our experiments, the images follow the 256-color storage convention, where pixel values occupy 8-bit storage space. Given the storage budget of images per class (IPC), the maximum storage budget for one class is capped at $8 \times \text{IPC} \times CHW$, where $C, H \text{ and } W$ represent the channel, height and width of the images, respectively.
When representing a colorful image pixel value with n bits, where $1 \leq n < 8$, there can be at most $2^n$ distinct colors per image. This must satisfy the condition $\sum_{i=1}^{2^{8-n}} N_{i} \leq 2^8$, where $N_i$ is the number of colors for the i-th color reduced image and each $N_i \leq 2^n$. Therefore, for images with n-bit format, up to $2^{8-n}$ colors can be represented in the storage budget using a bitmap index with small bytes. The bitmap index indicates the image number associated with the current lower bit color value.

\begin{table}[t]
    \caption{Test accuracy (\%) of previous works and our method on ConvNet D3. Our synthetic images are reduced from 256 colors to 64 colors. Our method outperforms previous methods and achieves state-of-the-art performance. }
    \label{Parameterization Performance Table}
    \centering
    \resizebox{1\linewidth}{!}{
    \begin{tabular}{c|c|ccc|ccc}
         \toprule
         \multicolumn{2}{c}{Dataset} & \multicolumn{3}{c}{CIFAR10} & \multicolumn{3}{c}{CIFAR100} \\
         \cmidrule(r){1-2} \cmidrule(r){3-5} \cmidrule{6-8}
         \multicolumn{2}{c}{IPC} & 1 & 10 & 50 & 1 & 10 & 50 \\
         \cmidrule{1-2} \cmidrule{3-5} \cmidrule{6-8}
         \multirow{3}{*}{Coreset} & Random & 14.4$\pm$0.2 & 26.0$\pm$1.2 & 43.4$\pm$1.0 & 4.2$\pm$0.3 & 14.6$\pm$0.5 & 30.0$\pm$0.4 \\
         & Herding \cite{welling2009herding} & 21.5$\pm$1.3 & 31.6$\pm$0.7 & 40.4$\pm$0.6 & 8.4$\pm$0.3 & 17.3$\pm$0.3 & 33.7$\pm$0.5 \\
         & K-Center \cite{sener2017active} & 23.3$\pm$0.9 & 36.4$\pm$0.6 & 48.7$\pm$0.3 & 8.6$\pm$0.3 & 20.7$\pm$0.2 & 33.6$\pm$0.4 \\
         \midrule
         \multirow{4}{*}{Distillation} & DD \cite{wang2018dataset} & - & 36.8$\pm$1.2 & - & - & - & - \\
         & DM \cite{zhao2023dataset} & 26.0$\pm$0.8 & 48.9$\pm$0.6 & 63.0$\pm$0.4 & 11.4$\pm$0.3 & 29.7$\pm$0.3 & 43.6$\pm$0.4 \\
         & DC \cite{zhao2020dataset} & 28.3$\pm$0.5 & 44.9$\pm$0.5 & 53.9$\pm$0.5 & 12.8$\pm$0.3 & 25.2$\pm$0.3 & - \\
         & TM \cite{cazenavette2022dataset} & 46.3$\pm$0.8 & 65.3$\pm$0.7 & 71.6$\pm$0.2 & 24.3$\pm$0.3 & 40.1$\pm$0.4 & 47.7$\pm$0.2 \\
         & DATM \cite{guo2023towards} & 46.9$\pm$0.5 & 66.8$\pm$0.2 & 76.1$\pm$0.3 & 27.9$\pm$0.2 & 47.2$\pm$0.4 & \textbf{55.0}$\pm$0.2 \\
         \midrule
         \multirow{6}{*}{Parameterization} & IDC \cite{kim2022dataset} & 50.0$\pm$0.4 & 67.5$\pm$0.5 & 74.5$\pm$0.1 & - & - & - \\
         & HaBa \cite{liu2022dataset} & 48.3$\pm$0.8 & 48.3$\pm$0.8 & 48.3$\pm$0.8 & 33.4$\pm$0.4 & 40.2$\pm$0.2 & 47.0$\pm$0.2 \\
         & RTP \cite{deng2022remember} & \textbf{66.4}$\pm$0.4 & 71.2$\pm$0.4 & 73.6$\pm$0.5 & 34.4$\pm$0.4 & 42.9$\pm$0.7 & - \\
         & SPEED \cite{wei2024sparse} & 63.2$\pm$0.1 & 73.5$\pm$0.2 & 77.7$\pm$0.4 & \textbf{40.0}$\pm$0.4 & 45.9$\pm$0.3 & 49.1$\pm$0.2 \\
         & FReD \cite{shin2024frequency} & 60.6$\pm$0.8 & 70.3$\pm$0.3 & 75.8$\pm$0.1 & 34.6$\pm$0.4 & 42.7$\pm$0.2 & 47.8$\pm$0.1 \\
          & \cellcolor{lightgray!30} AutoPalette & \cellcolor{lightgray!30} 58.6$\pm$1.1 & \cellcolor{lightgray!30} \textbf{74.3}$\pm0.2$ & \cellcolor{lightgray!30}  \textbf{79.4}$\pm0.2$ & \cellcolor{lightgray!30} 38.0$\pm0.1$ & \cellcolor{lightgray!30} \textbf{52.6}$\pm0.3$ & \cellcolor{lightgray!30} 53.3$\pm$0.8 \\
         \bottomrule
    \end{tabular}
    }
\end{table}

\section{Experiments}
In this section, we first evaluate the effectiveness of our method in comparison with other parameterization methods on various datasets. Afterwards, we perform experiments on the relations between synthetic image color number and model performance. We also conduct ablation studies and assess the efficacy of each proposed component to distillation performance.

\subsection{Experimental Setting}
We conduct experiments of our model on various benchmark datasets, including CIFAR-10 \cite{krizhevsky2009learning}, CIFAR-100 \cite{krizhevsky2009learning} and ImageNet \cite{deng2009imagenet}. We compare our parameterization method with core-set methods and other existing DD works containing baselines such as DD \cite{wang2018dataset}, DM \cite{zhao2023dataset}, DC \cite{zhao2020dataset}, TM \cite{cazenavette2022dataset}, and parameterization techniques including IDC \cite{kim2022dataset}, HaBa \cite{liu2022dataset}, RTP \cite{deng2022remember}, SPEED \cite{wei2024sparse}, FReD \cite{shin2024frequency}. Experiments are performed on different distillation memory budget settings for 1/10/50 images per class (IPC). We follow the previous works to use a ConvNetD3 for the CIFAR family and ConvNetD5 for ImageNet as the training and evaluation network. We follow the DATM \cite{guo2023towards} implementation based on trajectory matching, without soft label initialization using correctly predicted samples. Each experiment is evaluated on 5 randomly initialized networks, and the mean and standard deviation of the evaluation accuracy are recorded.  We set loss coefficients $\alpha$=1, $\beta$=1, $\gamma$=3 for all experiments if not specified. All experiments can be conducted on 2$\times$Nvidia H100 GPUs that have 80GB RAM for each or 4$\times$Nvidia V100 GPUs that have 32GB RAM for each.

\subsection{Experimental Results}
\textbf{Results on CIFAR10 and CIFAR100.} We perform experiments under paramterization settings on CIFAR10 \cite{krizhevsky2009learning}, CIFAR100 \cite{krizhevsky2009learning}. We set color palette network to condense the number of colors of a single image from 256 to 64, such that despite huge color space reduction quantized images still preserve much fidelity of images. As shown in Table \ref{Parameterization Performance Table}, our method achieves superior performance than other parameterization works in various tasks. Notably, in the experiments when IPC equals 10 and 50, our method significantly outperforms other methods. In CIFAR100 experiments, our model achieves 52.6\% and 53.3\% classification accuracy when IPC is respectively 10 and 50, which increases 6.7\% and 4.2\% higher than previous state-of-the-art parameterization methods. These outstanding performances highlight that reducing the color redundancy within the synthetic dataset can improve the storage utility and thereby improve the distillation result.

\textbf{Results on ImageNet.}
Following \cite{cazenavette2022dataset}, we conduct experiments on six subsets of ImageNet, where each subset consists of 10 classes and the images are of resolution 128$\times$128. We conduct experiments with the storage budget of IPC=10. ConvNetD5 is employed as the backbone model for training and evaluation. From Table \ref{tab:imagenet}, we can see our method outperforms other PDD methods on most of ImageNet subsets, including ImageNette, ImageWoof, ImageMeow and ImageYellow, while results of the other subsets still achieve comparable performance with previous state-of-the-art results. Specifically, our method achieves $44.3\%$ and $72.0\%$ on hard datasets ImageWoof and ImageYellow, increasing $0.2\%$ and $1.5\%$ than previous best PDD methods. We also observe that for subsets with distinct classes, our method achieves promising results, which is because color condensation effectively preserves the key semantics necessary for accurate classification. On the other hand, for fine-grained subsets where the classes are similar, we observe inferior performance. This is likely because fine-grained details are blurred in images represented by fewer colors, thereby making it challenging to differentiate between   classes.

\begin{table}[!t]
    \caption{Test accuracy (\%) on ImageNet-Subset: ImageNette, ImageWoof, ImageFruit, ImageMeow, ImageSquawk, ImageYellow. All experiments are conducted on CIFAR10 with IPC=10 storage budget for parameterization methods.}
    \label{tab:imagenet}
    \centering
    \resizebox{1\linewidth}{!}{
    \begin{tabular}{c|c|c|c|c|c|c}
        \toprule
        \multicolumn{1}{c}{Dataset} & \multicolumn{1}{c}{ImageNette} & \multicolumn{1}{c}{ImageWoof} & \multicolumn{1}{c}{ImageFruit} & \multicolumn{1}{c}{ImageMeow} & \multicolumn{1}{c}{ImageSquawk} & \multicolumn{1}{c}{ImageYellow} \\
        \midrule
        TM \cite{cazenavette2022dataset} & 63.0$\pm$1.3 & 35.8$\pm$1.8 & 40.3$\pm$1.3 & 40.4$\pm$2.2 & 52.3$\pm$1.0 & 60.0$\pm$1.5 \\
        HaBa \cite{liu2022dataset} & 64.7$\pm$1.6 & 38.6$\pm$1.3 & 42.5$\pm$1.6 & 42.9$\pm$0.9 & 56.8$\pm$1.0 & 63.0$\pm$1.6 \\
        FrePo \cite{shin2024frequency} & 66.5$\pm$0.8 & 42.2$\pm$0.9 & - & - & - & - \\
        SPEED \cite{wei2024sparse} & 72.9$\pm$1.5 & 44.1$\pm$1.4 & \textbf{50.0}$\pm$0.8 & 52.0$\pm$1.3 & \textbf{71.8}$\pm$1.3 & 70.5$\pm$1.5 \\
        \cellcolor{lightgray!30} AutoPalette & \cellcolor{lightgray!30} \textbf{73.2}$\pm$0.6 & \cellcolor{lightgray!30} \textbf{44.3}$\pm$0.9 & \cellcolor{lightgray!30}48.4$\pm$1.8 & \cellcolor{lightgray!30} \textbf{53.6}$\pm$0.7 & \cellcolor{lightgray!30} 68.0$\pm$1.4 & \cellcolor{lightgray!30} \textbf{72.0}$\pm$1.6 \\
        \bottomrule
    \end{tabular}
    }
    
\end{table}

\begin{table}[!b]
    \begin{minipage}{.48\linewidth}
      \caption{Test accuracy (\%) when a certain loss component is removed during training.}
      \centering
        \begin{tabular}{*{3}{c}|c}
             \toprule
             $\mathcal{L}_{m}$ & $\mathcal{L}_{b}$ & $\mathcal{L}_{a}$ & Accuracy \\
             \midrule
             \xmark & & & 64.00 \\
             \midrule
              & \xmark & & 61.40 \\
             \midrule
              & & \xmark & 60.14 \\
             \midrule
             \checkmark & \checkmark & \checkmark & \textbf{66.20} \\
             \bottomrule
        \end{tabular}
        \label{tab:loss_effect}
    \end{minipage}%
    \hfill
    \begin{minipage}{.48\linewidth}
      \centering
        \caption{Evaluation on the effectiveness of submodular selection using quantized images, in comparison with random initialization and submodular selection using full color images.}
        \begin{tabular}{*{2}{c}}
            \toprule
            Initialization Method & Accuracy \\
            \midrule
            Random Real  & 60.84 \\
            Graph Cut Real & 61.41 \\
            Graph Cut on Quantized Image & \textbf{62.13} \\
            \bottomrule
        \end{tabular}
        \label{Exp: Submodular on quantized}
    \end{minipage} 
\end{table}

\textbf{Compatibility of Distillation Frameworks.}
While we take trajectory matching as our primary distillation method, we demonstrate that our framework can effortlessly be equipped to improve other dataset distillation methods. As illustrated in Table \ref{table: different frameworks}, our method shows a significant performance boost across all IPC settings and datasets when adopted to the distribution matching method. Especially, our method increases the test accuracy up to $15\%$ when IPC=10, and $15.7\%$ for IPC=1 on CIFAR100. This observation aligns with our objectives that our color-oriented redundancy management framework should be adapted across different standard DD frameworks The performance improvement underscores the high compatibility of our methods with diverse DD frameworks.

\begin{table}[!t]
    \caption{Test accuracy (\%) on different DD frameworks including DM and TM across various IPC settings. We adopt ConvNetD3 as the backbone network to distil the synthetic dataset on CIFAR10 and CIFAR100.}
    \label{table: different frameworks}
    \centering
    \resizebox{.85\linewidth}{!}{
    \begin{tabular}{c|c|ccc|ccc}
         \toprule
         \multicolumn{2}{c}{Framework} & \multicolumn{3}{c}{DM \cite{zhao2023dataset}} & \multicolumn{3}{c}{TM \cite{cazenavette2022dataset}} \\
         \midrule
         Dataset & IPC & Vanilla & AutoPalette & Increase & Vanilla & AutoPalette & Increase \\
         \midrule
         \multirow{3}{*}{CIFAR10} & 1 & 26.0$\pm$0.8 & 35.5$\pm$0.4 & 9.5$\uparrow$& 46.3$\pm$0.8 & 58.6$\pm$1.1 & 12.3$\uparrow$ \\
         & 10 & 48.9$\pm$0.6 & 60.9$\pm$0.1 & 12.0$\uparrow$ & 65.3$\pm$0.7 & 74.3$\pm$0.2 & 9.0$\uparrow$ \\
         & 50 & 63.0$\pm$0.4 & 71.6$\pm$0.4 & 8.6$\uparrow$ & 71.6$\pm$0.2 & 79.4$\pm$0.2 & 7.8$\uparrow$ \\

         \midrule
         \multirow{3}{*}{CIFAR100} & 1 & 11.4$\pm$0.3 & 20.9$\pm$0.1 & 9.5$\uparrow$ & 24.3$\pm$0.3 & 38.0$\pm$0.1 & 15.7$\uparrow$ \\
         & 10 & 29.7$\pm$0.3 & 44.7$\pm$0.1 & 15.0$\uparrow$ & 40.1$\pm$0.4 & 52.6$\pm$0.3 & 12.5$\uparrow$ \\
         & 50 & 43.6$\pm$0.4 & 50.1$\pm$0.1 & 6.5$\uparrow$ & 47.7$\pm$0.2 & 54.1$\pm$0.8 & 5.6$\uparrow$ \\
         \bottomrule
    \end{tabular}
    }
\end{table}

\subsection{Ablation Study}
Here, we focus on comparing variants of our proposed framework. Therefore, we fix the number of synthetic images to 10 per class, rather than fully utilizing the available storage capacity.


    


\textbf{Effectiveness of Loss Components.} To validate the contribution of each loss term to the overall framework, we conduct experiments on CIFAR10 with IPC=10. Specifically, we construct three variants of our model by removing $\mathcal{L}_m$, $\mathcal{L}_{b}$, and $\mathcal{L}_{a}$, correspondingly. We show the experimental result in Table \ref{tab:loss_effect}. Under the same experimental conditions, eliminating specific loss functions will suffer from performance decline. The results demonstrate the essential role played by each loss function in optimizing the palette network.

\textbf{Effectiveness of Selection Criteria in the Color-guided Initialization.} We compare our proposed initialization with two baseline sample selection criteria, including Random Real, and Graph Cut Real. Random real is widely adopted by DD methods, which randomly select images from the full dataset as the initialization of the synthetic dataset. In Graph Cut Real, we apply graph cut on 8-bit images and select the most representative samples with high information gain. Compared with Graph Cut Real, the selection criteria used in our framework computes the information gain over the quantized images. From table \ref{Exp: Submodular on quantized}, we can see that our methods using quantized images exhibit better performance than comparison approaches. The graph cut with original full-color images also outperforms the baseline model using 
randomly selected real images as initialization, confirming the effectiveness of\begin{wrapfigure}{r}{0.4\textwidth}
    \centering
        \includegraphics[width=0.4\textwidth]{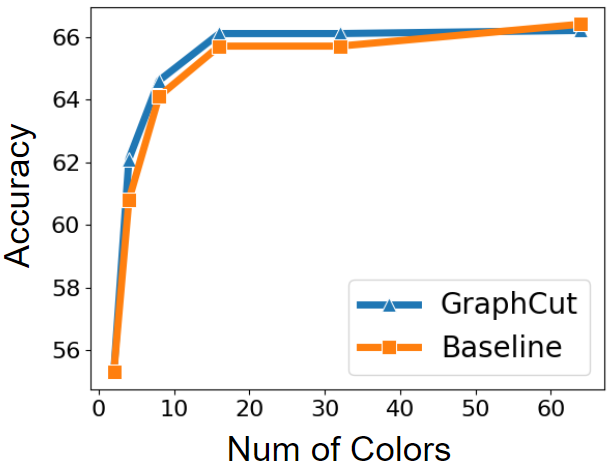}
    \caption{Comparison between the performance of submodular color diversity initialization and random real images initialization.}
    \label{fig:init_effect}
\end{wrapfigure} 
computing information gain over color-reduced images

\textbf{Effectiveness of Color-guided Initialization under Different Color Depth.}
Our experiments compare the performance of two ways to initialize the base images: one employing our method of color-guided initialization (denotes GraphCut), and randomly selecting real images as the base images (denotes Baseline). We contrast two approaches on CIFAR10 with IPC=10, spanning from synthesizing images with low-bit quantization to those with higher-bit quantization. From Figure \ref{fig:init_effect}, we observe that our method brings better performance when quantized images are represented in lower bits. Starting from 16 colors per pixel, we observe a gradual convergence in performance as we move towards utilizing full-color space. Meanwhile, as can be seen, when $K$=32 and 64, it achieves the best trade-off, as it still attains competitive performance comparable to that of full-color budget images, while substantially reducing the storage space required for quantized images.

\section{Conclusions, Limitations, and Future Work}
In this paper, we aim to solve the color-redundancy issue within data distillation from both the image level and the dataset level. For condensing the colors within images, we utilize a palette network to capture the color redundancy between pixels and represent images using fewer colors. Beyond this, to reduce repetitive patterns in between synthetic images design a guided image initialization module that selects samples by maximising information gain. Extensive experimental results demonstrate that our AutoPalette framework can effectively reduce color redundancy and simultaneously preserve the essential low-level feature for model training.

\textbf{Limitations.} For instance, images from different classes may have a bias towards color usage. Images of one class may be sufficient to be represented by fewer colors than those from other classes, in which case an imbalanced color budget arrangement may be a better option. In future, it is also promising to explore the dynamic color depth allocation, which allocates more budgets to difficult classes, thereby improving the distillation performance.

\begin{ack}
This research is partially supported by the Australian Research Council (DE240100105, DP240101814, DP230101196)
\end{ack}

\bibliographystyle{plain}
\bibliography{neurips_2024.bib}

\newpage

\appendix\section{Appendix}

\subsection{Proof for Conditional Information Gain of Graph Cut}
\label{proof:graph cut}
The generalized graph cut set function is defined as:
\begin{equation}
\label{graphcut definition}
    f(\mathcal{A}) = \lambda \sum_{i \in \mathcal{T}} \sum_{a \in \mathcal{A}} Sim(i, a) - \sum_{a_1, a_2 \in \mathcal{A}}  Sim(a_1, a_2),
\end{equation}
where $\mathcal{A} \subset \mathcal{T}$, and $Sim$ is a similarity function.
The submodular conditional gain is defined as:
\begin{equation}
\label{conditional gain definition}
    f(\mathcal{A}|\mathcal{B}) \triangleq f(\mathcal{A} \cup \mathcal{B}) - f(\mathcal{B}),
\end{equation}
which indicates the gain by adding samples in set $\mathcal{B}$ to $\mathcal{A}$. By substituting Eq. \eqref{conditional gain definition} with \eqref{graphcut definition}, we can obtain:
\begin{equation}
\begin{split}
    f(\mathcal{A}|\mathcal{B}) = & \lambda \sum_{i \in \mathcal{T}} \sum_{c \in \mathcal{A} \cup \mathcal{B}} Sim(i, c) - \sum_{c_1, c_2 \in \mathcal{A} \cup \mathcal{B}} Sim(c_1, c_2) \\
    & - \lambda \sum_{i \in \mathcal{T}} \sum_{b \in \mathcal{B} } Sim(i, b) + \sum_{b_1, b_2 \in \mathcal{B}} Sim(b_1, b_2).
\end{split}
\end{equation}





In the sample selection cases, $\mathcal{A}$ and $\mathcal{B}$ are disjoint, and thus $\sum_{c \in \mathcal{A} \cup \mathcal{B}}$ can be rewritten as $\sum_{c \in \mathcal{A} - \mathcal{B}}$. Therefore, we can reformulate $f(\mathcal{A} \cup \mathcal{B})$:
\begin{equation}
\begin{split}
    f(\mathcal{A} \cup \mathcal{B}) = & \lambda \sum_{i \in \mathcal{T}}\sum_{c \in \mathcal{A} \setminus \mathcal{B}} Sim(i, c) + \lambda \sum_{i \in \mathcal{T}} \sum_{b \in \mathcal{B}} Sim(i, b) \\
    & - \sum_{b_1, b_2 \in \mathcal{B}} Sim(b_1, b_2) - \sum_{c_1, c_2 \in \mathcal{A} \setminus \mathcal{B}} Sim(c_1, c_2) \\
    & - 2\sum_{a' \in \mathcal{A} \setminus \mathcal{B}} \sum_{b \in \mathcal{B}} Sim(a', b).
\end{split}
\end{equation}

We can then formulate Eq. \eqref{conditional gain definition} as:
\begin{equation}
    f(\mathcal{A}|\mathcal{B}) = f(\mathcal{A} \setminus \mathcal{B}) - 2\sum_{a' \in \mathcal{A} \setminus \mathcal{B}} \sum_{b \in \mathcal{B}} Sim(a', b).
\end{equation}

When $\mathcal{A}$ and $\mathcal{B}$ are disjoint, $\mathcal{A}$ is independent of $\mathcal{B}$ and we then simplify the conditional gain of graph cut function to:
\begin{equation}
    f(\mathcal{A}|\mathcal{B}) = f(\mathcal{A}) - 2\sum_{a' \in \mathcal{A} \setminus \mathcal{B}} \sum_{b \in \mathcal{B}} Sim(a', b).
\end{equation}

\subsection{Experimental Details}

\textbf{Datasets.}
We conduct experiments on multiple datasets:
\begin{itemize}
    \item CIFAR10: an image dataset consists of 50,000 32$\times$32 RGB images for training, and 10,000 images for testing. CIFAR10 contains 10 classes: airplane, automobile, bird, cat, deer, dog, frog, horse, ship, truck.
    \item CIFAR100: an image dataset that is similar to CIFAR100, but has 100 classes containing 600 images each.
    \item ImageNet subsets: high resolution image subsets of ImageNet \cite{deng2009imagenet} that contain all 128$\times$128 RGB images and each contains 10 classes. ImageNet subsets include include ImageNette, ImageWoof, ImageFruit, ImageMeow, ImageSquawk, and ImageYellow.
\end{itemize}

\textbf{Networks.}
For the experiments of low resolution datasets including CIFAR10 and CIFAR100, 3-layer convolutional neural networks (ConvNet) are employed and we follow the identical network structures to the previous works. Each convolution layer contains 128 3$\times$3 filters, followed by an instance normalization layer, a ReLU, and an average pooling layer with 2$\times$2 kernel and stride 2. For the high resolution datasets such as ImageNet subsets, we use 5-layer ConvNets to perform the experiments. For the cross architecture experiments, we employ VGG11 \cite{karen2014very}, AlexNet \cite{krizhevsky2012imagenet}, and RestNet18 \cite{he2016deep} and follow the implementations of the previous DD works. 
The palette network consists of two convolution layers and one ReLU between them. Both two convolution layers have 1$\times$1 kernels and the second layer has no bias.

\textbf{Implementation details.}
While we primarily use Trajectory matching (TM) as the distillation objectives, our method can be seamlessly adapted into other DD frameworks for various downstream tasks and datasets. In Table \ref{tab:hyper dm} and \ref{tab:hyper tm}, we provide the hyper-parameters settings in our work for both TM and DM on different datasets. Specifically, although our method is insensitive for most of the hyper-parameters, certain parameters including synthetic steps, the maximum starting epoch and the synthetic batch size should be carefully examined.

\subsection{Algorithm for Sample Selection in Initialization}
    \begin{algorithm}[!t]
    \caption{Algorithm for guided image selection with maximum information gain.}
    \label{algorihtm: submodular}
    \Input{Original Dataset $\mathcal{T}$; Number of classes $N_{class}$; Images per class \textit{IPC}; Network $\theta$}
    \Output{Set of selected data samples $\mathcal{A}$ for each class}
    \FOR{$i \leftarrow 1$ \KwTo $N_{class}$}{
        $X^{i} \gets $  all images belong to class i from $\mathcal{T}$\; 
        $\mathcal{A}^{i} \gets \{rand(X^{i})\}$ \tcp*{Initialize the selected set with a random sample}
        
        \FOR{$j \leftarrow 2$ \KwTo IPC}{
            $\mathcal{C} \gets \mathcal{T}^{i} \setminus \mathcal{A}^{i}$ \;
            $c^{*} \gets $ Eq. \eqref{eq: selection criteria} \tcp*{Select the data sample with the highest gain}
            $\mathcal{A}^{i} \gets \mathcal{A}^{i} \cup \{c^{*}\}$ \;
        }
    }
    
\end{algorithm}

\subsection{Cross Architecture Performance}
One of the main concerns in data distillation arises from synthetic dataset overfitting to the training models, resulting in limited generalizability to be used by the other network architectures. Therefore, cross-architecture performance is crucial for evaluating the effectiveness of data distillation methods. To assess the generalized performance of our method, we employ CIFAR10 synthetic datasets trained on ConvNet to train various network structures including VGG11 \cite{karen2014very}, AlexNet \cite{krizhevsky2012imagenet} and ReNnet18 \cite{he2016deep}. The results of cross-network architecture are presented in Table \ref{tab:cross_arch}. We noticed that when IPC is relatively higher, our method outperforms the other baseline methods.

\begin{table}[ht]
    \caption{Cross architecture performance of synthetic dataset that is optimized by ConvNet. Experiments are performed on CIFAR10, and VGG11, AlexNet and ResNet18 are used for evaluating the cross architecture performance.}
    \label{tab:cross_arch}
    \centering
    \resizebox{1\linewidth}{!}{
    \begin{tabular}{c|ccc|ccc|ccc}
    
         \toprule
         & \multicolumn{3}{c}{VGG11} & \multicolumn{3}{c}{AlexNet} & \multicolumn{3}{c}{ResNet18} \\
         \midrule
         \diagbox[width=\dimexpr \textwidth/8+2\tabcolsep\relax, height=1cm]{Method}{IPC} & 2 & 11 & 51 & 2 & 11 & 51 & 2 & 11 & 51 \\
         \midrule
         TM \cite{cazenavette2022dataset} & 38.0 $\pm$ 1.2 & 50.5 $\pm$ 1.0 & 61.4 $\pm$ 0.3 & 26.1 $\pm$ 1.0 & 36.0 $\pm$ 1.5 & 49.2 $\pm$ 1.3 & 35.2 $\pm$ 1.0 & 45.1 $\pm$ 1.5 & 54.5 $\pm$ 1.0 \\
         IDC \cite{kim2022dataset} & 48.2 $\pm$ 1.2 & 52.7 $\pm$ 0.7 & 65.2 $\pm$ 0.6 & 32.5 $\pm$ 2.2 & 43.7 $\pm$ 3.0 & 54.9 $\pm$ 1.1 & 46.7 $\pm$ 0.9 & 50.2 $\pm$ 0.6 & 64.5 $\pm$ 1.2 \\
         HaBa \cite{liu2022dataset} & 48.3 $\pm$ 0.5 & \textbf{60.5 $\pm$ 0.6} & 67.5 $\pm$ 0.4 & 43.6 $\pm$ 1.5 & 49.0 $\pm$ 3.0 & 60.1 $\pm$ 1.4 & 47.4 $\pm$ 0.7 & 58.0 $\pm$ 0.9 & 64.4 $\pm$ 0.6 \\
         FReD \cite{shin2024frequency} & \textbf{50.1 $\pm$ 0.8} & 60.0 $\pm$ 0.6 & 69.9 $\pm$ 0.4 & \textbf{44.1 $\pm$ 1.3} & \textbf{55.9} $\pm$ 0.8 & 65.9 $\pm$ 0.8 & \textbf{53.9 $\pm$ 0.7} & 64.4 $\pm$ 0.6 & 71.4 $\pm$ 0.7 \\
         \cellcolor{lightgray!30} AutoPalette & \cellcolor{lightgray!30} 41.3 $\pm$ 1.1 & \cellcolor{lightgray!30} 57.6 $\pm$ 1.1 & \cellcolor{lightgray!30} \textbf{70.3 $\pm$ 0.2} & \cellcolor{lightgray!30} 36.7 $\pm$ 2.5 & \cellcolor{lightgray!30} 44.5 $\pm$ 1.1 & \cellcolor{lightgray!30} \textbf{72.5 $\pm$ 0.2} & \cellcolor{lightgray!30} 46.7 $\pm$ 1.2 & \cellcolor{lightgray!30} \textbf{66.0 $\pm$ 1.3} & \cellcolor{lightgray!30} \textbf{75.8 $\pm$ 0.2} \\
    
         \bottomrule
    \end{tabular}
    }
    
\end{table}

\subsection{Number of colors vs Performance}
Our method still demonstrates competitive performance, even when synthetic images are quantized into extremely low bits, as shown in Table \ref{Exp: performance vs quality}. The test results for CIFAR10 IPC=10 is provided, from which we can observe a slight performance gap when the number of colors used for quantized images decreases to merely 8 colors per pixel (3 bits) from the original 256 colors per pixel (8 bits). The results demonstrate that our color quantization method captures key features for synthetic data and appropriately condenses images into fewer colors.

\begin{table}[ht]
    \centering
    \caption{Model performance when evaluated on the synthetic dataset using at most different number of colors.}
    \resizebox{.7\linewidth}{!}{
    \begin{tabular}{c|ccccccc}
        \toprule
        \#colors & 256 (full) & 64 & 32 & 16 & 8 & 4 & 2 \\
        \midrule
        Accuracy & 66.8 & 66.2 & 66.1 & 65.9 & 64.6 & 62.1 & 55.3 \\
        \bottomrule
    \end{tabular}
    }
    \label{Exp: performance vs quality}
\end{table}

\subsection{Traditional color Quantization Methods}
Our palette network aims to identify the essential color characteristics of the synthetic dataset and represent it with fewer colors. To test the efficacy of the palette network, we compare it with other color quantization methods on DD tasks, including Median Cut \cite{heckbert1982color} and OCTree \cite{gervautz1988simple}. We conduct the experiments on CIFAR10, and Table \ref{tab:traditional quantization comparison} shows that the palette network achieves the superior performances under different IPC settings.


\begin{table}[!t]
    \centering
    \caption{Test accuracy (\%) in comparison with the traditional color quantization methods. Quantization methods are applied to synthetic dataset for CIFAR10, when IPC=10 and 50.}
    \resizebox{0.7\textwidth}{!}{
    \begin{tabular}{c|cccccc}
        \toprule
        \multicolumn{1}{c}{IPC} &  \multicolumn{6}{c}{10} \\
        \midrule
        \multicolumn{1}{c}{Scale} & 2 & 4 & 8 & 16 & 32 & 64 \\
        \midrule
        Median Cut & 42.5 & 46.4 & 52.4 & 55.6 & 57.6 & 60.7 \\
        OCTree & 23.7 & 30.7 & 41 & 53.1 & 59.6 & 60.8 \\
        \cellcolor{lightgray!30} Palette Network & \cellcolor{lightgray!30} \textbf{55.3} & \cellcolor{lightgray!30} \textbf{62.1} & \cellcolor{lightgray!30} \textbf{64.6} & \cellcolor{lightgray!30} \textbf{65.9} & \cellcolor{lightgray!30} \textbf{66.1} & \cellcolor{lightgray!30} \textbf{66.2} \\
        \bottomrule
    \end{tabular}
    }
    \label{tab:traditional quantization comparison}
\end{table}


\begin{table}[!t]
    \centering
    \caption{Hyperparameters for our method based on Distribution matching (DM) framework.}
    \begin{tabular}{c|c|c|c|c|c}
    \toprule
    Dataset & IPC & Synthetic batch size & Palette network LR & Synthetic Image LR & ZCA \\
    \midrule
    \multirow{3}{*}{CIFAR10} & 1 & - & 0.05 & 1 & True \\
    & 10 & - & 0.05 & 1 & True \\
    & 50 & - & 0.05 & 10 & True \\
    \midrule
    \multirow{3}{*}{CIFAR100} & 1 & - & 0.05 & 1 & True \\
    & 10 & - & 0.05 & 1 & True \\
    & 50 & 50 & 0.05 & 10 & True \\

    \bottomrule
    \end{tabular}
    \label{tab:hyper dm}
\end{table}

\begin{table}[!th]
    \caption{Hyperparameters for our method based on Trajectory matching (TM) framework.}
    \centering
    \resizebox{1\linewidth}{!}{
    \begin{tabular}{c|c|c|c|c|c|c|c|c|c|c}
    \toprule
    Dataset & IPC & Synthetic batch size & Synthetic steps & Expert epochs & Max start epoch & Palette network LR & Synthetic Image LR & Step size LR & Teacher LR & ZCA \\
    \midrule
    \multirow{3}{*}{CIFAR10} & 1 & - & 80 & 2 & 15 & 0.05 & 500 & $10^{-7}$ & $10^{-2}$ & True \\
    & 10 & - & 35 & 2 & 40 & 0.05 & 1000 & $10^{-5}$ & $10^{-2}$ & True \\
    & 50 & 600 & 40 & 2 & 50 & 0.05 & 500 & $10^{-5}$ & $10^{-2}$ & True \\
    \midrule
    \multirow{3}{*}{CIFAR100} & 1 & - & 60 & 2 & 20 & 0.05 & 1000 & $10^{-5}$ & $10^{-2}$ & True \\
    & 10 & 600 & 35 & 2 & 70 & 0.05 & 1000 & $10^{-5}$ & $10^{-2}$ & True \\
    & 50 & 200 & 60 & 2 & 70 & 0.05 & 1000 & $10^{-5}$ & $10^{-2}$ & True \\
    \midrule
    \multirow{1}{*}{ImageNette} & 10 & 80 & 20 & 2 & 20 & 0.01 & 10000 & $10^{-4}$ & $10^{-2}$ & False \\
    
    \bottomrule
    \end{tabular}
    }
    
    \label{tab:hyper tm}
\end{table}

\clearpage
\newpage
\subsection{Visualization of Distilled Images}
\begin{figure}[!ht]
    \centering
    \includegraphics[width=\textwidth]{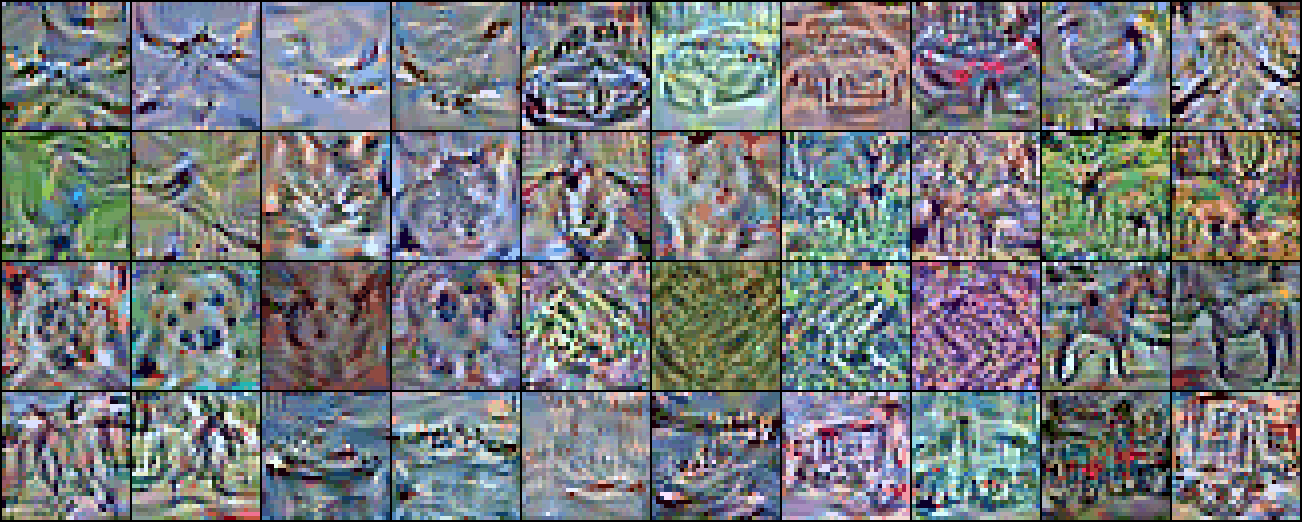}
    \caption{CIFAR10 color condensed synthetic images with ZCA whitening.}
    \label{fig:cifar10 64 zca}
\end{figure}

\begin{figure}[!ht]
    \centering
    \includegraphics[width=\textwidth]{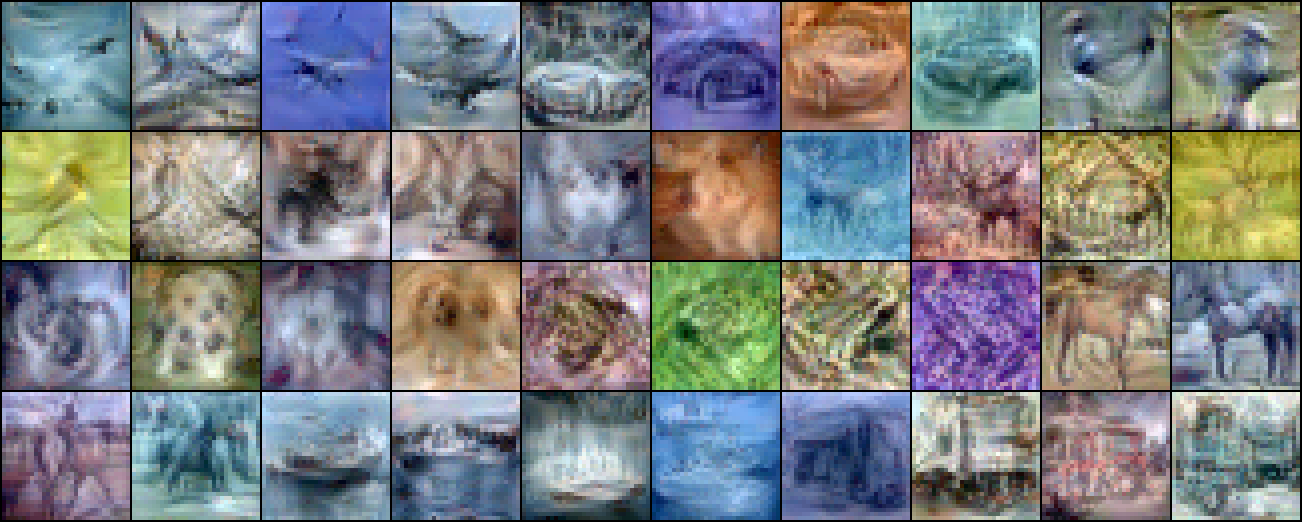}
    \caption{CIFAR10 color condensed synthetic images without ZCA whitening.}
    \label{fig:cifar10 64 no zca}
\end{figure}

\begin{figure}[!ht]
    \centering
    \includegraphics[width=\textwidth]{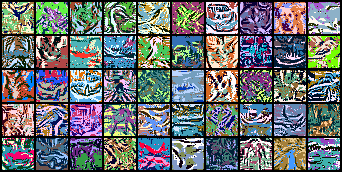}
    \caption{CIFAR10 synthetic images in 3-bit color depth}
    \label{fig:cifar10 low bit}
\end{figure}

\clearpage
\newpage
\begin{figure}[!ht]
    \centering
    \includegraphics[width=0.8\textwidth]{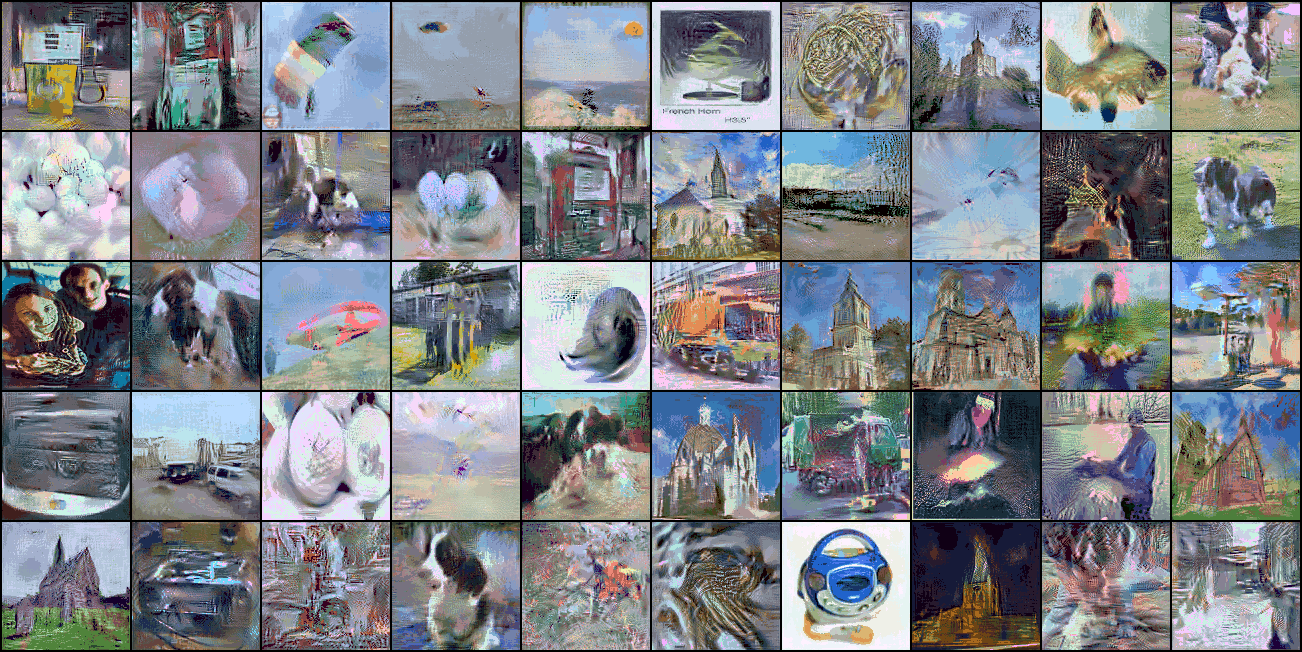}
    \caption{Color condensed synthetic images for ImageNette}
    \label{fig:imagenette}
\end{figure}

\begin{figure}[!ht]
    \centering
    \includegraphics[width=0.8\textwidth]{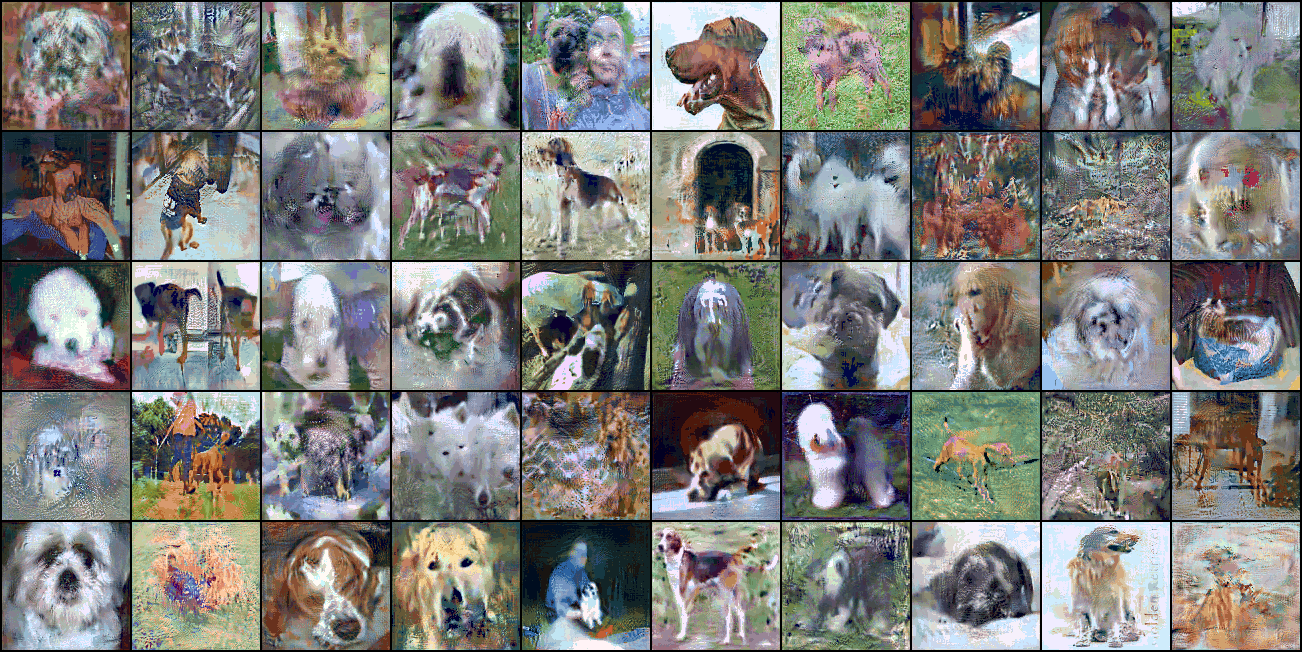}
    \caption{Color condensed synthetic images for ImageWoof}
    \label{fig:imagewoof}
\end{figure}

\begin{figure}[!ht]
    \centering
    \includegraphics[width=0.8\textwidth]{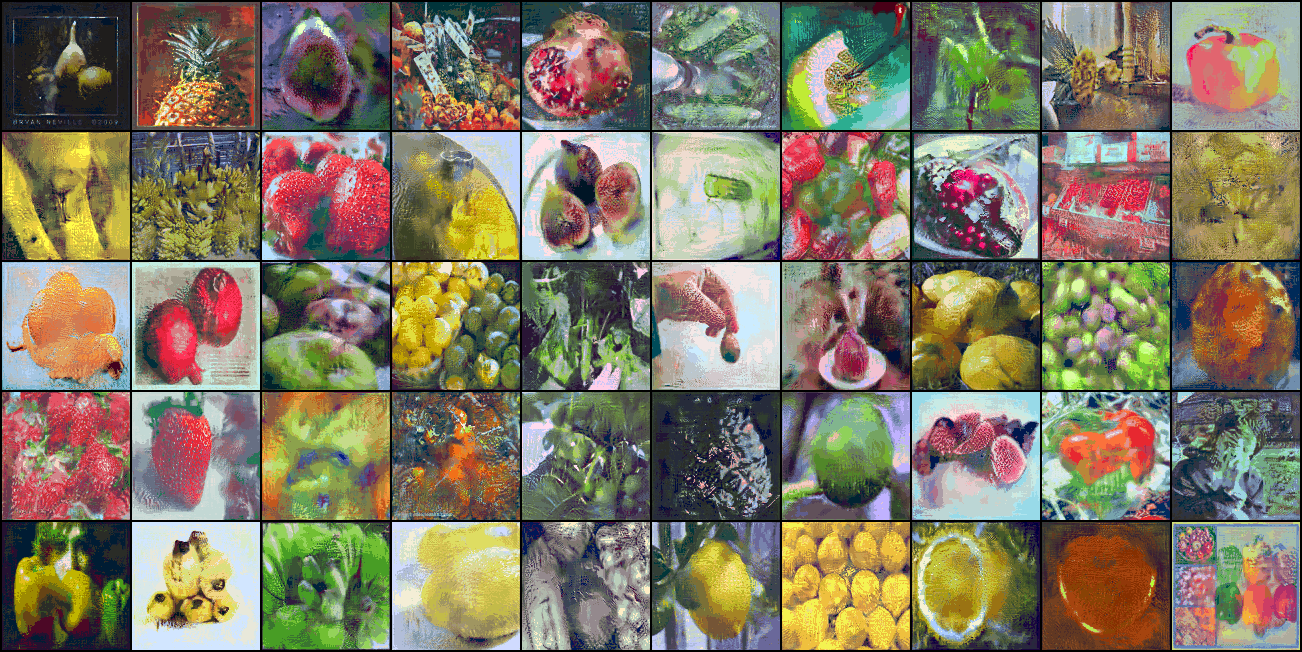}
    \caption{Color condensed synthetic images for ImageFruit}
    \label{fig:imagefruit}
\end{figure}

\begin{figure}[!ht]
    \centering
    \includegraphics[width=0.8\textwidth]{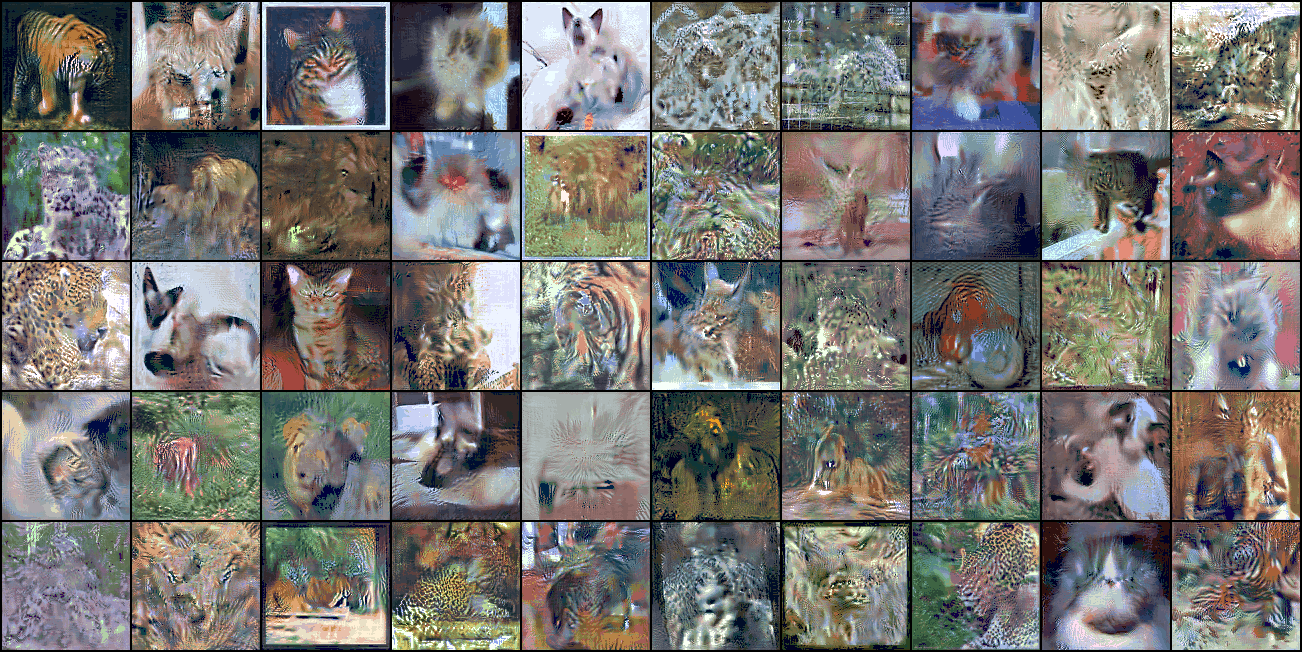}
    \caption{Color condensed synthetic images for ImageMeow}
    \label{fig:imagemeow}
\end{figure}

\begin{figure}[!ht]
    \centering
    \includegraphics[width=0.8\textwidth]{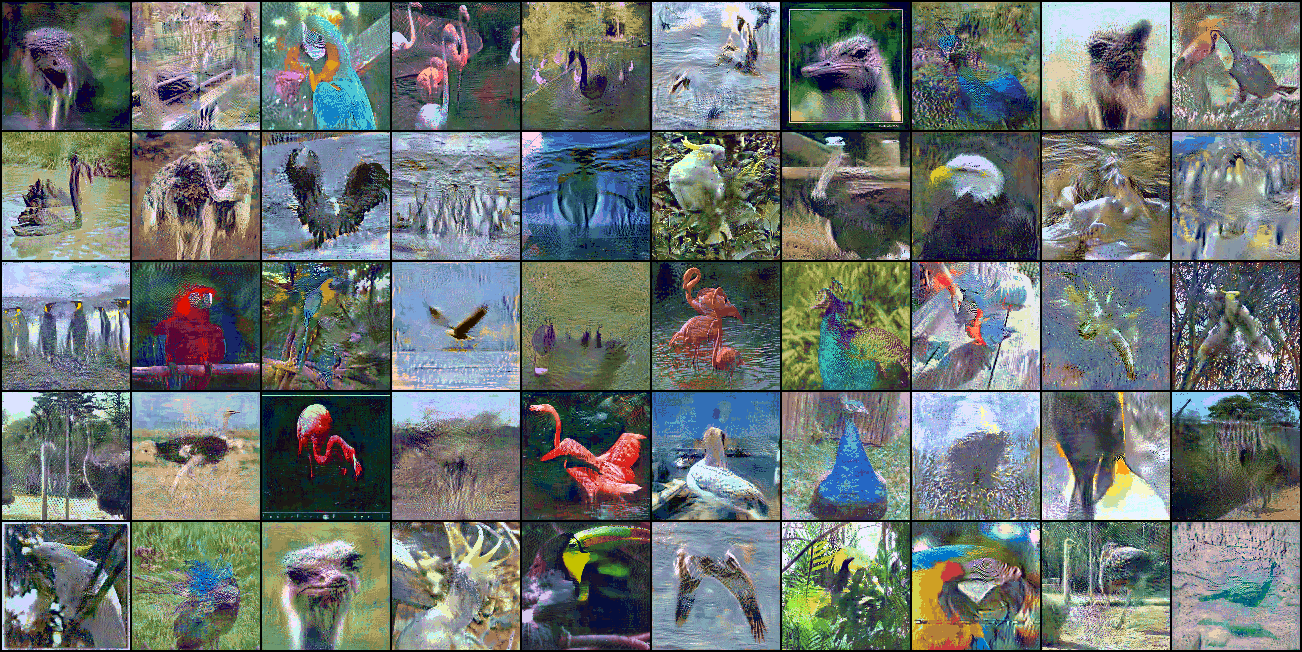}
    \caption{Color condensed synthetic images for ImageSquawk}
    \label{fig:imagesquawk}
\end{figure}

\begin{figure}[!ht]
    \centering
    \includegraphics[width=0.8\textwidth]{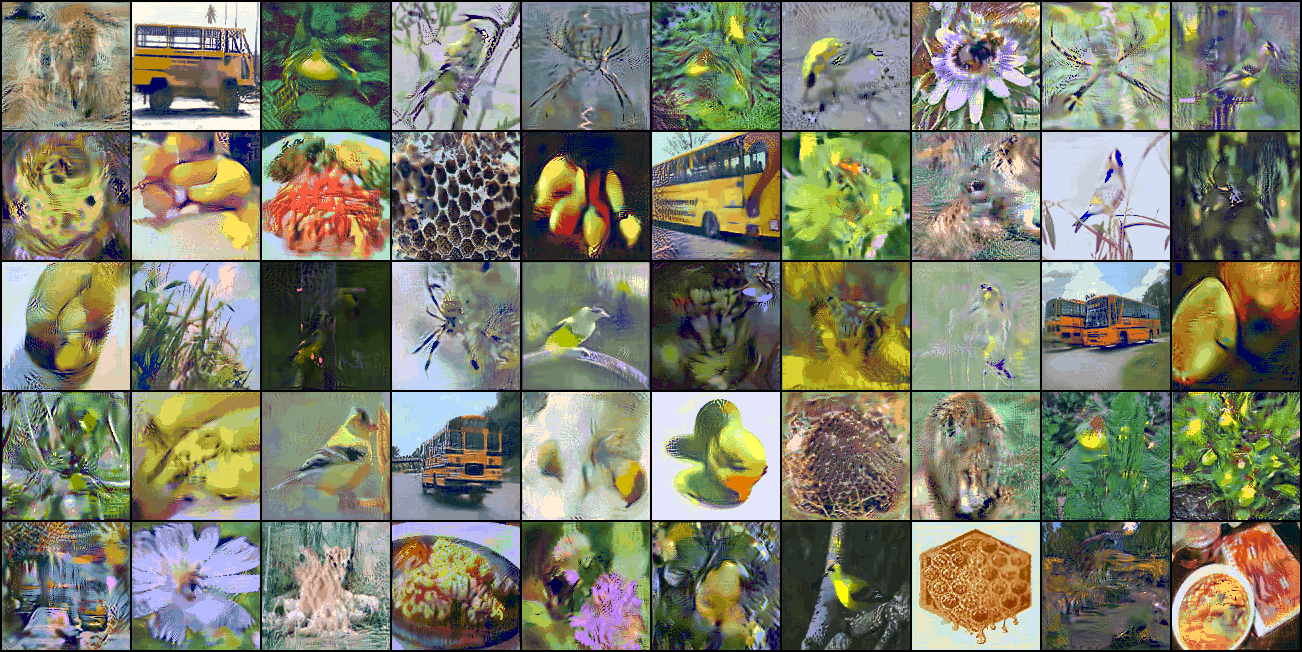}
    \caption{Color condensed synthetic images for ImageYellow}
    \label{fig:imageyellow}
\end{figure}

\end{document}